# Particle swarm optimization model to predict scour depth around bridge pier

Shahaboddin Shamshirband [1,2], Amir Mosavi [3,4], Timon Rabczuk [5]

[1] Department for Management of Science and Technology Development, Ton Duc Thang University, Ho Chi Minh City, Vietnam; shahaboddin.shamshirband@tdtu.edu.vn

[2] Faculty of Information Technology, Ton Duc Thang University, Ho Chi Minh City, Vietnam

[3] Institute of Automation, Kando Kalman Faculty of Electrical Engineering, Obuda University, Budapest-1034, Hungary; amir.mosavi@kvk.uni-obuda.hu

[4] School of the Built Environment, Oxford Brookes University, Oxford-OX30BP, United Kingdom; a.mosavi@brookes.ac.uk

[5] Institute of Structural Mechanics, Bauhaus University Weimar, 99423 Weimar, Germany; timon.rabczuk@uni-weimar.de

**Abstract:** Scour depth around bridge piers plays a vital role in the safety and stability of the bridges. Existing methods to predict scour depth are mainly based on regression models or black box models in which the first one lacks enough accuracy while the later one does not provide a clear mathematical expression to easily employ it for other situations or cases. Therefore, this paper aims to develop new equations using particle swarm optimization as a metaheuristic approach to predict scour depth around bridge piers. To improve the efficiency of the proposed model, individual equations are derived for laboratory and field data. Moreover, sensitivity analysis is conducted to achieve the most effective parameters in the estimation of scour depth fo r both experimental and filed data sets. Comparing the results of the proposed model with those of existing regression-based equations reveal the superiority of the proposed method in terms of accuracy and uncertainty. Moreover, the ratio of pier width to flow depth and ratio of d50 (mean particle diameter) to flow depth for the laboratory and field data were recognized as the most effective parameters, respectively. The derived equations can be used as a suitable proxy to estimate scour depth in both experimental and prototype scales.

1. Introduction

Scour depth around pier foundation plays an important role in safety and operation of bridges as an element of infrastructures. Due to complex mechanism of flow around piers, it is difficult to estimate scour depth appropriately. Sound prediction of scour depth is of great interest for hydraulic and bridge engineers to design a safe structure. Moreover, as the bridge is aging, signs of weakening due to scour depth around piers are expected to be highlighted and even though it can exceed until a complete failure. Therefore, development of predictive models with enough accuracy and reliability should be taken under consideration by engineers.

Generally, methods applied for prediction of local scour depth around pier foundation can be categorized as black box models and empirical equations. Black box models such as artificial neural networks, support vector regression models find relationship between as set of effective variables as the model inputs and the scour depth as target variable. In this approach, there is no need to introduce the mathematical relationships between input and output variables. They have been frequently used to estimate scour depth around bridge piers using different sample sizes and for different situations (Azamathulla and Ghani, 2010; Zounemat-Kermani et al., 2009). On the contrary, many empirical equations based on linear or nonlinear models were proposed using principal variables influencing the scouring phenomenon (Johnson, 1992; Melville and Chiew, 1999; Richardson and Davis, 2001). These models were implemented using either laboratory data or filed data. Bateni et al. (2007) applied artificial neural network (ANN) and adaptive neuro-fuzzy inference system (ANFIS) models to estimate scour depth around bridge piers. In this regard, numerous laboratory data were employed to train and validate efficiency of the proposed models. The results demonstrated that the models have a good capability to estimate the target variable. Using extensive datasets of field measurement, Azamathulla et al. (2009) and Pal et al. (2012) derived new equations based on genetic programming and M5 model, respectively. They found that the proposed equations perform equally well or even better than ANN models. Liao et al. (2015) applied a probabilistic criterion to evaluate potential of pier scour in a river basin in Taiwan. The results indicated efficiency of the proposed method for stability and reliability analysis. Recently, Sharafi et al. (2016) developed two predictive models including support vector regression as a black box model and nonlinear regression based equation for estimation of scour depth for real field dataset. The results revealed superiority of black box model over regression based equations. The black box models may be capable to find complex and nonlinear relationship between input and output variables and consequently providing more accurate predictions. However, the models based on empirical equations in comparison with black box methods have two main advantages;1) they are transparent and offer more insight of the physics of the problem, 2) their easier implementation (Alizadeh et al., 2017).

Training the data driven models with different data (data selection and number of data sample) can affect the models' performance remarkably (Tinoco et al., 2015). Thus, the models developed using different datasets and techniques may provide different estimations of scour depth for different conditions of flows and waterways. Therefore, the models developed using laboratory data may provide inaccurate or uncertain predictions of scour depth for real field bridges. Mohamed et al. (2006) explored validation of some available formulae for scour depth estimations. In a separate study, Johnson et al. (2015) quantified parameter uncertainty and model uncertainty in commonly used scour equations. The results indicated that for practical applications, uncertainty in model and parameters should be addressed accordingly.

The main objective of this study is therefore to employ an evolutionary based algorithm to improve accuracy and uncertainty of scour depth predictions. Two equations were derived to estimate scour depth for laboratory and field data individually. To find the most effective parameters on the target variable, sensitivity analysis for both datasets were conducted separately. In section 2, data resources, materials and methods are briefly descried. Results related to sensitivity analysis, accuracy and uncertainty of estimations obtained from the proposed equations are discussed in section 3. Concluding remarks are presented in the last section.

2. Materials and methods
    2.1. Data resources

In this study, an attempt was made to employ an extensive available datasets to cover a wide variation of hydraulic and geometric characteristics. Moreover, previous studies were mainly based only on laboratory dataset or only on field measurement while in this study both datasets are explored. Thus, total number of 552 laboratory data and 540 data of field measurement were used for the model development. The filed and laboratory data measured by several researchers and organizations were presented by Benedict and Caldwell (2014). Statistical analysis of the applied datasets is given in Table 1 where "Min", "Max', "Mean" and "Std" stands for minimum, maximum, average, and standard deviation of the variable. In Tale 1, D is pier width normal to the flow, V flow mean velocity; Vc is sediment critical velocity, L is pier length, y is approach flow depth, d50 is Grain size where 50 percent of the bed material is finer by weight and $\sigma$ is sediment gradation. The scour depth and Froude number are also represented with S and Fr, respectively. The rest are dimensionless form of the variables.

Table 1. Statistical analysis of laboratory and field datasets

|  | Laboratory | | | | Field | | | |
|---|---|---|---|---|---|---|---|---|
|  | Min | Max | Mean | Std | Min | Max | Mean | Std |
| D(m) | 0.01585 | 0.915 | 0.107 | 0.1435 | 0.3048 | 28.7 | 2.797 | 3.674 |
| V(m/s) | 0.149 | 2.16 | 0.512 | 0.3216 | 0.088 | 4.084 | 1.366 | 0.75 |
| $V_c$(m/s)-L(m) | 0.222 | 1.27 | 0.443 | 0.2213 | 0.975 | 38.1 | 10.705 | 4.255 |
| y(m) | 0.0201 | 1.9 | 0.269 | 0.2513 | 0.1524 | 22.524 | 4.163 | 3.518 |
| d50(m) | 0.00022 | 0.0078 | 0.00118 | 1.37 | 0.000008 | 0.108 | 0.01675 | 25.15 |
| S(m) | 0.00399 | 1.41 | 0.1357 | 0.149 | 0 | 10.393 | 1.0528 | 1.404 |
| $\sigma$ | 1.1 | 5.5 | 1.454 | 0.692 | 1.2 | 20.34 | 3.358 | 2.806 |
| $V/V_c$-L/y | 0.4148 | 5.38 | 1.254 | 0.846 | 0.5142 | 81.818 | 5.3487 | 6.575 |
| D/y | 0.0477 | 19.16 | 0.704 | 1.724 | 0.0722 | 50.297 | 1.251 | 2.805 |
| d50/y | 0.00012 | 0.107 | 0.007 | 0.0107 | 4.9e-7 | 0.2264 | 0.0106 | 0.0226 |
| Fr | 0.067 | 1.498 | 0.377 | 0.248 | 0.0269 | 1.184 | 0.2703 | 0.177 |
| S/y | 0.0198 | 6.87 | 0.754 | 0.805 | 0 | 3.4 | 0.3066 | 0.2948 |

Due to different geometry and flow conditions for different rivers, Table 1 shows that sediment gradation, D/y, and d50/y changes in a wider range for filed data compared with laboratory data. This difference may add flow complexity and its interaction with pier foundation. Consequently, the derived equations based on laboratory data may not describe real scour depth process around real world bridge piers accordingly. Therefore, efficiency of the existing equations as well as development of new equations based on field measurements should be explored.

### 2.2. Particle swarm optimization algorithm

Recently, evolutionary algorithm as a generic population based metaheuristic optimization algorithm attracted attention of many researchers in different fields of study. They have been widely used for a large number of engineering applications to optimize design parameters. Particle swarm optimization (PSO) and genetic algorithm (GA) are two types of population based computation commonly used for optimization problems. However, PSO algorithm has advantages of faster convergence and easier implementation than GA. PSO algorithm firstly developed by Eberhart and Kennedy (1995) mimics social behavior of bird flocking in search of food. Each bird called particle represent a single solution in search space. In brief, the algorithm initializes with randomly generated particles searching for optimum solution. Each particle has a position and velocity. Using two values of "$P_s$" and "$G$", generations will be obtained through an iterative process. "$P_s$" represents best solution of each particle it has achieved so far and "$G$" is the best value obtained so far by any particle in the population. Dealing with PSO, there are some tunable parameters including number of particles, neighborhood size, number of iterations, acceleration coefficient ($C_1, C_2$) and inertia weight ($w$) in which they have to be determined carefully in order to achieve desirable performance. At each time step, the algorithm works by pushing each particle toward its "$P_s$" and "$G$" locations. Following equations (1) and (2), position and velocity of each particle in space is updated as:

$$V_s(t+1) = wV_s(t) + C_1 r_1 (P_s - X_s(t)) + C_2 r_2 (G - X_s(t)) \tag{1}$$

$$X_s(t+1) = X_s(t) + V_s(t+1) \tag{2}$$

where $V_s(t)$ represents the velocity vector of particle $s$ in $t$ time; $X_s(t)$ represents the position vector of particle $s$ in $t$ time; $r_1$ and $r_2$ are two random functions in the range [0, 1]. The best solution is the solution with the least error or the nearest one to the target. Therefore, the optimization algorithm tries to achieve the best solution by minimization of the error. Detailed description of the algorithm can be found in Eberhart and Kennedy (1995). Figure 1 provides a schematic layout of the PSO algorithm.

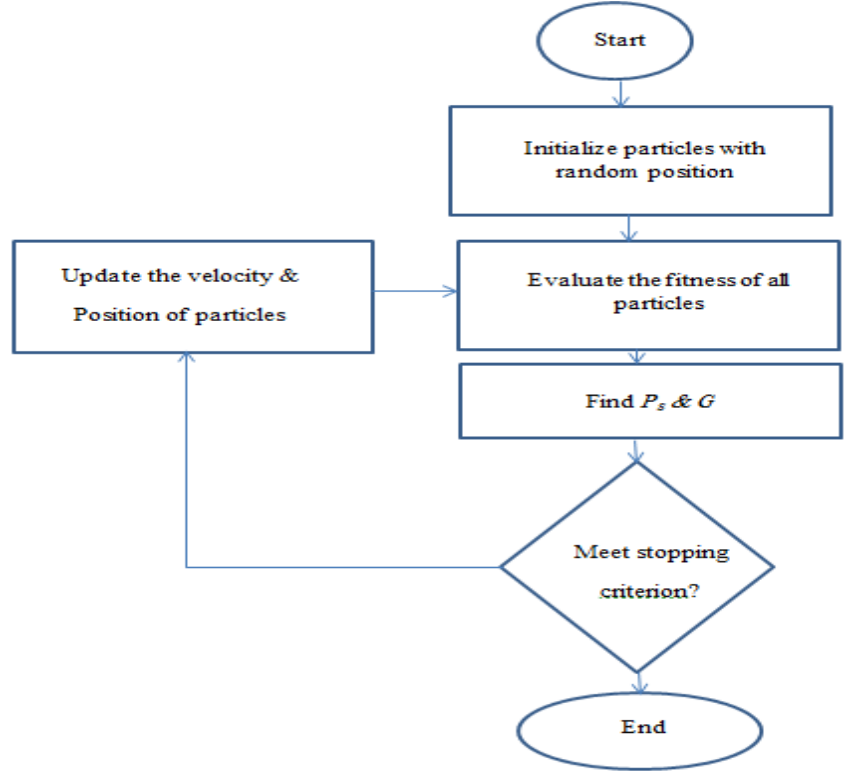

Figure 1. Schematic layout of PSO algorithm

2.3. Existing equations

To estimate scour depth around bridge piers, many investigators developed different models using laboratory/field data. Regardless of black box models, several techniques based on regression or optimization algorithm were applied to derive mathematical expressions for scour depth estimation. Dimensional analysis is a common task prior to developing empirical equations in which main parameters influencing the target variable are taken under consideration. Assuming a circular pier in a steady flow, main components affecting scour depth in a laboratory condition can be stated as:

$$S = f(\rho, \mu, \sigma, V, g, y, d50, V_c, D) \tag{3}$$

where $\rho, \mu, g$ are the fluid density, the fluid dynamic viscosity, and the gravitational acceleration, respectively. Similarly, for prototype environment, the following parameters are considered as the main component in estimation of scour depth:

$$S = f(\rho, \mu, \sigma, V, g, y, d50, L, D) \tag{4}$$

Formerly, these variables have been defined in subsection 2.1. This study does not present details of dimensional analysis and more information on the topic can be found from literature. However, by summarizing the effective variables, equations (5) and (6) are obtained to describe scour depth in laboratory and field scales.

$$\frac{S}{y} = f(\sigma, Fr, \frac{D}{y}, \frac{d50}{y}, \frac{V}{V_c}) \qquad (5)$$

$$\frac{S}{y} = f(\sigma, Fr, \frac{D}{y}, \frac{d50}{y}, \frac{L}{y}) \qquad (6)$$

The only difference in equations (5) and (6) indicates that pier length is more important for prototype environment while in laboratory investigation it is more common to consider ratio of flow velocity to sediment critical velocity. In this regard, many empirical equations using different statistical and optimization techniques were derived using different datasets. Table 2 presents some of these equations used for scour depth estimations.

Table 2. Empirical equations for estimation of scour depth around bridge piers

| Model | Equation | Type of dataset |
|---|---|---|
| Laursen and Toch (1956) | $S/y = 1.35(D/y)^{0.7}$ | Laboratory |
| Shen et al. (1969) | $S/y = 3.4(Fr)^{0.67}(D/y)^{0.67}$ | Laboratory |
| Hancu (1971) | $S/D = 2.42(2\frac{V}{V_c} - 1)(\frac{V_c}{gD})^{1/3}$ | Laboratory |
| Melville and Sutherland (1988) | $S/D = K_I K_D K_y K_\alpha K_S$ *** | Laboratory |
| Johnson (1992) | $S/y = 2.02(\sigma)^{-0.98}(Fr)^{0.21}(D/y)^{0.98}$ | Laboratory |
| Richardson and Davis (2001) | $S/y = 2.6(Fr)^{0.65}(D/y)^{0.43}$ | Field |
| HEC-18 (Mohamed et al., 2005) | $S/y = 2.1(Fr)^{0.43}(D/y)^{0.65}$ | Laboratory and Field |
| Azamathulla et al. (2009) | $S/y = 1.82(\sigma)^{-0.03159}(Fr)^{0.42}(d50/y)^{0.042}(D/y)^{-0.28}(L/y)^{-0.37}$ | Field |
| Sharafi et al. (2016) | $S/y = 0.28(\sigma)^{0.13}(Fr)^{0.47}(d50/y)^{-0.1}(D/y)^{0.44}(L/y)^{0.23}$ | Field |

*** $K_I, K_D, K_y, K_\alpha,$ and $K_S$ are the flow intensity, sediment size, flow depth, pier shape, and alignment coefficients, respectively.

Following Table 2 it can be found that effective variables have different values when they have been derived using different datasets and techniques. For example, relative scour depth increases with increase in $D/y$ and $L/y$ for model proposed by Sharafi et al. (2016) whereas an inverse relationship for these variables with scour depth can be found when the model of Azamathulla et al. (2009) is used. Thus, it can be obtained that the relationship is strongly to data availability and also the applied method. Finally, it is noticed that the coefficient of variables for laboratory and field data differ to some extend even though the models were developed in dimensionless form. Dealing with empirical equations, model simplicity, data availability and computational effort is also of great importance. Moreover, consistency of the relationship with physics of the problem should be taken under consideration.

### 2.4. Model application and evaluation

Considering available datasets, different combinations of effective variables can be employed for model development. Moreover, the model uncertainty and its sensitivity to each variable can provide useful information in scour depth studies. Moreover, suitability of derived equations for laboratory or field data is a key issue toward obtaining reliable estimations. In this regard, an extensive datasets of both laboratory and field measurements were collected from literature. Sensitivity analysis for both datasets was conducted individually. In other words, several models including different combinations of input variables were employed in the model development. Performance of each model was evaluated using error measures to find effect of each variable on the scour depth estimation. Finally, efficiency of derived equations for laboratory and field data is investigated and also compared against performance of existing equations. It should be noticed that 70% of the datasets were used for model development and 30% remained to test and compare efficiency of the proposed equations. Considering equations (5) and (6), several models using different combinations of effective parameters are employed to conduct sensitivity analysis for dimensionless parameter of scour depth. Generally, equations (7) and (8) are considered as the basic expression to estimate dimensionless parameter of scour depth in laboratory and field scales.

$$\frac{S}{y} = a(\sigma)^b (Fr)^c \left(\frac{D}{y}\right)^d \left(\frac{d50}{y}\right)^e \left(\frac{V}{V_c}\right)^f \tag{7}$$

$$\frac{S}{y} = a(\sigma)^b (Fr)^c \left(\frac{D}{y}\right)^d \left(\frac{d50}{y}\right)^e \left(\frac{L}{y}\right)^f \tag{8}$$

where $a, b, c, d, e, f$ are coefficients of the input variables. These coefficients are obtained using PSO algorithm in a way the best value of objective function is achieved. Using PSO algorithm, different functions can be set as the objective function and also multiple objective functions can employed as well. In this study, the objective function was defined to minimize the root mean square error of estimations. Therefore, different models were developed to investigate efficiency of the effective parameters. Table 3 describes effective variables for laboratory and filed data where *L* and *F* represent models for laboratory and field estimations, respectively. In Table 3, the target variable for all models is $\frac{S}{y}$.

Table 3. Model specifications for sensitivity analysis of scour depth

| Model No. | Input variables | Model No. | Input variables |
|---|---|---|---|
| L1 | $\sigma, Fr, \frac{D}{y}, \frac{d50}{y}, \frac{V}{V_c}$ | F1 | $\sigma, Fr, \frac{D}{y}, \frac{d50}{y}, \frac{L}{y}$ |

| | | | | |
|---|---|---|---|---|
| L2 | $Fr, \dfrac{D}{y}, \dfrac{d50}{y}, \dfrac{V}{V_c}$ | F2 | $Fr, \dfrac{D}{y}, \dfrac{d50}{y}, \dfrac{L}{y}$ | |
| L3 | $\sigma, \dfrac{D}{y}, \dfrac{d50}{y}, \dfrac{V}{V_c}$ | F3 | $\sigma, Fr, \dfrac{D}{y}, \dfrac{d50}{y}$ | |
| L4 | $\sigma, Fr, \dfrac{d50}{y}, \dfrac{V}{V_c}$ | F4 | $\sigma, Fr, \dfrac{d50}{y}, \dfrac{L}{y}$ | |
| L5 | $\sigma, Fr, \dfrac{D}{y}, \dfrac{V}{V_c}$ | F5 | $\sigma, Fr, \dfrac{D}{y}, \dfrac{L}{y}$ | |
| L6 | $\sigma, Fr, \dfrac{D}{y}, \dfrac{d50}{y}$ | F6 | $\sigma, \dfrac{D}{y}, \dfrac{d50}{y}, \dfrac{L}{y}$ | |

To evaluate and compare performance of the PSO based models against those of existing equations, error measures including bias ($\bar{e}$), coefficient of determination ($R^2$), root mean square error (RMSE), and mean absolute error (MAE) were employed. Moreover, an extra index named $1.96 S_e$ was used to compute width of uncertainty band for the 95% confidence level. These indices are defined as follows:

$$\bar{e} = \frac{1}{n}\sum_{i=1}^{n} e_i \ ; \ e_i = y'_i - y_i \tag{9}$$

$$R^2 = 1 - \frac{\sum_{i=1}^{n}(y_i - y'_i)^2}{\sum_{i=1}^{n}(y_i - \bar{y})^2} \tag{10}$$

$$RMSE = \sqrt{\frac{\sum_{i=1}^{n}(y_i - y'_i)^2}{n}} \tag{11}$$

$$MAE = \frac{\sum_{i=1}^{n}|y'_i - y_i|}{n} \tag{12}$$

$$S_e = \sqrt{\sum_{i=1}^{n}(e_i - \bar{e})^2 / n - 1} \tag{13}$$

where $n$ is number of data sample, $y'_i$ is estimated value and $y_i$ is measured value of the target variable.

3. Results and discussion
   3.1. Laboratory data

Recognizing effective parameters and their influence on the scour depth is a key step to obtain a reliable estimation of the variable. In this regard, importance of each input variable on the target variable has been carried out using sensitivity analysis. To do that, PSO algorithm was applied to obtain the optimum coefficients for each case resulting in minimum error. Thus, the following equations excluding any effective variables can be considered as an optimum solution if the value of that specific input variable is not available. Error measures are used to evaluate performance of each model and effect of the input variables on the efficiency of the target variable. Table 4 gives mathematical expression including optimized values of coefficients and performance of the models.

Table 4. Derived equations and their performance for laboratory dataset during testing period

| Model No. | Derived equation | $R^2$ | RMSE (m) | $\bar{e}$ (m) | MAE (m) |
|---|---|---|---|---|---|
| L1 | $\dfrac{S}{y} = 1.282(\sigma)^{-0.397}(Fr)^{0.679}(\dfrac{D}{y})^{0.610}(\dfrac{d50}{y})^{-0.142}(\dfrac{V}{V_c})^{-0.476}$ | 0.877 | 0.046 | 0.009 | 0.029 |
| L2 | $\dfrac{S}{y} = 0.893(Fr)^{1.016}(\dfrac{D}{y})^{0.625}(\dfrac{d50}{y})^{-0.262}(\dfrac{V}{V_c})^{-0.836}$ | 0.861 | 0.049 | 0.008 | 0.030 |
| L3 | $\dfrac{S}{y} = 2.235(\sigma)^{-0.434}(\dfrac{D}{y})^{0.587}(\dfrac{d50}{y})^{0.111}(\dfrac{V}{V_c})^{0.212}$ | 0.897 | 0.043 | 0.010 | 0.029 |
| L4 | $\dfrac{S}{y} = 5(\sigma)^{-0.372}(Fr)^{-0.119}(\dfrac{d50}{y})^{0.354}(\dfrac{V}{V_c})^{0.305}$ | 0.119 | 0.161 | 0.059 | 0.098 |
| L5 | $\dfrac{S}{y} = 1.777(\sigma)^{-0.383}(Fr)^{0.323}(\dfrac{D}{y})^{0.595}(\dfrac{V}{V_c})^{-0.111}$ | 0.886 | 0.045 | 0.009 | 0.029 |
| L6 | $\dfrac{S}{y} = 1.874(\sigma)^{-0.421}(Fr)^{0.226}(\dfrac{D}{y})^{0.595}(\dfrac{d50}{y})^{0.029}$ | 0.891 | 0.044 | 0.009 | 0.029 |

According to Table 4 it can be concluded that generally all the models except *L4* provides relatively accurate estimations of scour depth for laboratory dataset. Exclusion of *D/y* degrade model's performance remarkably in which it can be considered as the most important parameter for estimation of the target variable. Estimations of the optimization model (*L4*) ignoring ratio of pier width to flow depth as input variable to shows low correlation with the measured values of the target variable. Moreover, high values of RMSE, MAE, and $\bar{e}$ confirms the model inefficiency. On the other hand, results of model *L6* reveal that excluding $\dfrac{V}{V_c}$ does not change the model performance significantly that it is interesting because measuring $V_c$ is not an easy task and its value is not available for many cases. However, it does overwhelm importance of flow velocity while it has been included as Froude number. Results of models *L3* and *L5* are comparable with model *L1* which show in these two models Froude number and $\dfrac{d50}{y}$ were excluded in model development respectively. Even though Froude number is excluded in model *L3*, its main components (*V* and *y*) have been considered at least once in other terms ($\dfrac{V}{V_c}, \dfrac{D}{y}, \dfrac{d50}{y}$). A comparison between results model *L1* and model *L2* demonstrates superiority of the model including $\sigma$ as the model input over the model excluding sediment gradation. Regarding the coefficients achieved from the optimization algorithms, an inverse relationship between $\sigma$ and *S* and a direct relationship between Froude number and *S* can be found in which are in agreement with the physics of the phenomenon. In other words, the finding show increasing trend in scour depth with increasing in Froude number or with decreasing in $\sigma$. Similarly, a direct relationship between scour depth and *D/y* implies with increasing in pier width, the scour depth is expected to increase. Considering coefficient for $\dfrac{V}{V_c}$ (negative values), it is not surprising that higher values of sediment critical velocity helps to lower scour depth around pier foundation. To provide more comparisons of performance of different developed model, an uncertainty analysis giving width of uncertainty band with 95% confidence level ($1.96S_e$) of each model is illustrated in Figure 2.

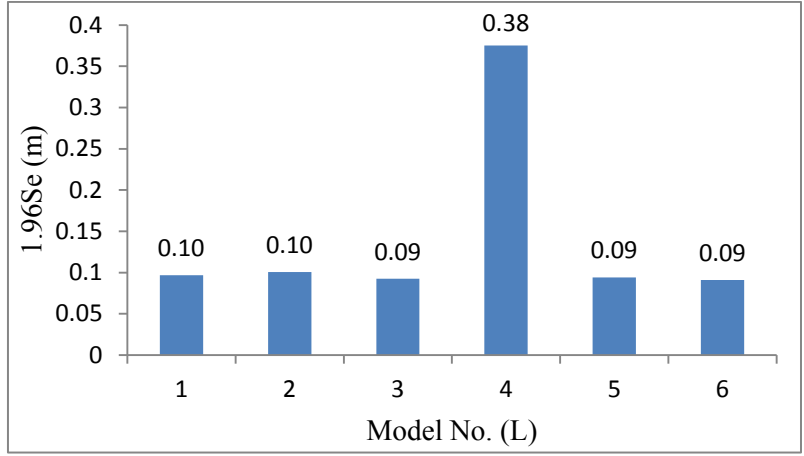

Figure 2. Width of uncertainty band for the derived equations (laboratory dataset)

As observed from Figure 2, width of uncertainty band for all the derived equations except *L4* is roughly equal. Results of uncertainty analysis for equations excluding effective variables are comparable with equation *L1*. Models *L3*, *L5*, and *L6* present the narrowest width of uncertainty slightly better performance than model *L1* and *L2* in terms of width of uncertainty band for scour depth estimation. Considering all the error evaluation criteria investigated in this study, models *L3* and *L6* are recognized as the best models to estimate scour depth for laboratory studies. However, to explore efficiency of the proposed equations, their performance is compared with those of existing equations. In this regard, the error measures were employed to evaluate performance of equations during testing period (for 30% of the datasets in which they were not included in the model development). Table 5 gives results of the best derived equations and also existing equations.

Table 5. Results of the best derived equations against existing equations

| Model | $R^2$ | RMSE (m) | $\bar{e}$ (m) | MAE (m) | $1.96 S_e$ (m) |
|---|---|---|---|---|---|
| Laursen and Toch (1956) | 0.894 | 0.061 | 0.032 | 0.0398 | 0.052 |
| Shen et al. (1969) | 0.616 | 0.105 | 0.059 | 0.069 | 0.087 |
| Hancu (1971) | 0.009 | 0.909 | 0.546 | 0.579 | 0.726 |
| Johnson (1992) | 0.677 | 0.082 | -0.011 | 0.037 | 0.081 |
| Richardson and Davis (2001) | 0.513 | 0.106 | 0.055 | 0.0708 | 0.091 |
| HEC-18 (Mohamed et al., 2005) | 0.829 | 0.057 | 0.025 | 0.038 | 0.051 |
| *L3 (this study)* | 0.897 | 0.043 | 0.010 | 0.029 | 0.042 |

Regarding Table 5, it can be concluded that the proposed equation (*L3*) outperforms existing equations for estimation of laboratory scour depth. It has the lowest values of RMSE, mean error, MAE and also the narrowest width of uncertainty band. Moreover, the highest correlation between scour depth estimations and laboratory measurements are obtained for the model *L3*. Regardless the proposed equation derived using PSO algorithm, equations of HEC-18 and Laursen and Toch (1956) are among the best models in terms of the error indices. Generally, results in Table 5 reveal that except the equation of Johnson (1992), the abovementioned equations overestimate scour depth in physical models. As many equations have been derived to estimate scour depth and analyze the results in dimensionless form, scatter plot for the best models of Table 5 is depicted to provide more comparisons of performance of the equations. Figure 3 depicts estimated dimensionless values of scour depth versus laboratory measurements for models with the best performance.

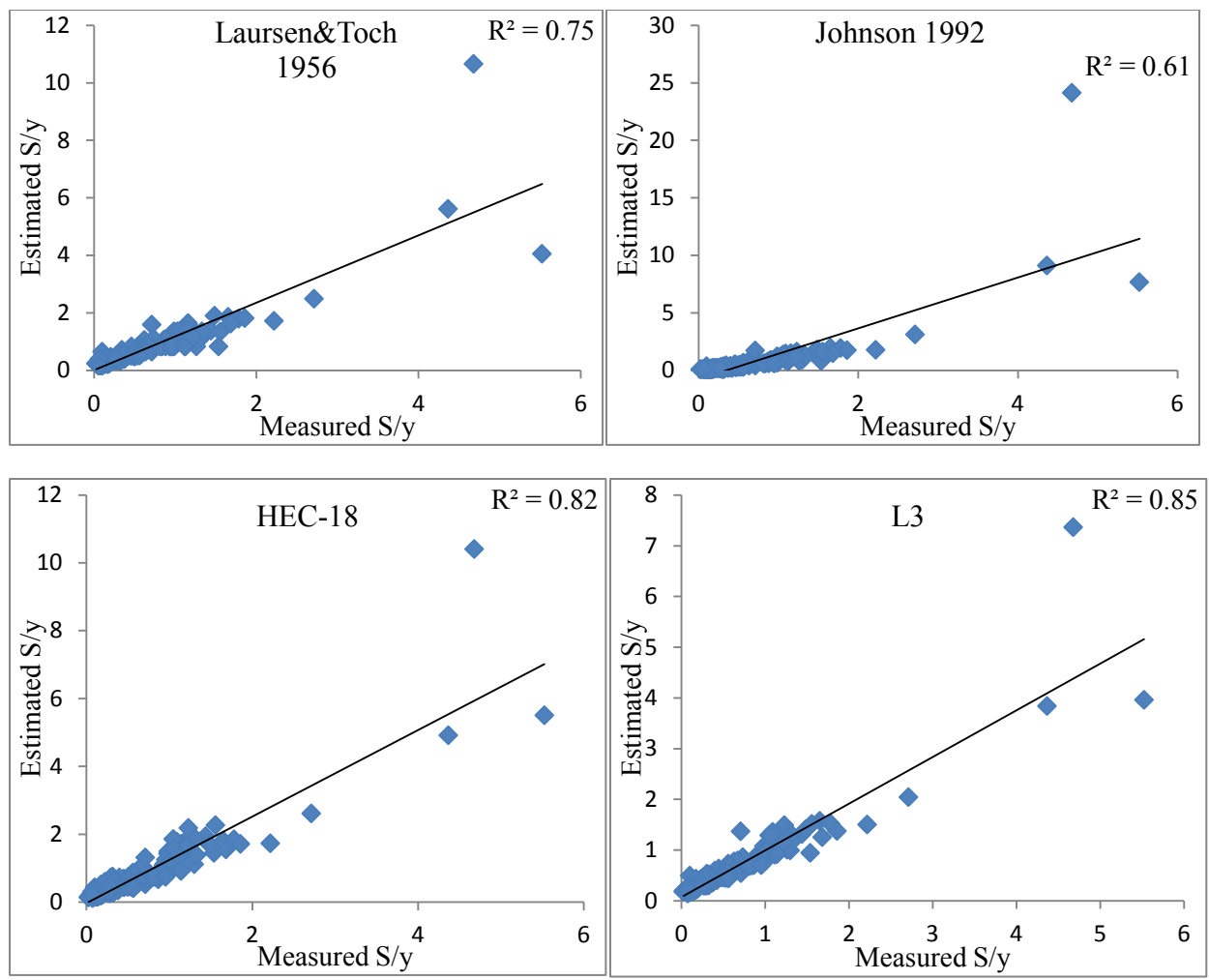

Figure 3. Scatter plot of estimated scour depth versus measured values

Figure 3 demonstrates superiority of estimations obtained from equation *L3* over the existing equations. Regarding the figure, a relatively high correlation between results of the derived equation and measured values can be observed. Moreover, the model has a good performance in estimation of high values of S/y while the other equations provide very conservative values or overestimate the values. The main purpose of developing models and deriving equations in dimensionless form is to generalize and extend their applicability for wider range or real world problems. In the following subsection, estimation of scour depth in prototype applications is discussed.

### 3.2. Field data

Reliable estimation of scour depth is important for engineering applications. Moreover, equations based on laboratory measurements may provide inaccurate estimations for filed applications. Prior to investigation of existing equations for prototype applications, sensitivity analysis was conducted to find the effective parameters on scour depth. In this regard, filed measurements obtained from literature are divided into two parts of calibration (70% of whole dataset) and testing (30% of remaining). Using PSO algorithm and general form of equation (8) for field data, optimized coefficients were computed for each case. Results of sensitivity analysis and derived equations for field dataset are given in Table 6.

Table 6. Derived equations and their performance for field dataset during testing period

| Model No. | Derived equation | $R^2$ | RMSE (m) | $\bar{e}$ (m) | MAE (m) |
|---|---|---|---|---|---|
| F1 | $\frac{S}{y} = 0.095 * (\sigma)^{0.116}(Fr)^{0.178}(\frac{D}{y})^{0.189}(\frac{d50}{y})^{-0.136}(\frac{L}{y})^{0.324}$ | 0.727 | 0.822 | 0.014 | 0.547 |
| F2 | $\frac{S}{y} = 0.1 * (Fr)^{0.154}(\frac{D}{y})^{0.219}(\frac{d50}{y})^{-0.145}(\frac{L}{y})^{0.308}$ | 0.774 | 0.753 | 0.030 | 0.520 |
| F3 | $\frac{S}{y} = 0.241 * (\sigma)^{0.075}(Fr)^{0.265}(\frac{D}{y})^{0.349}(\frac{d50}{y})^{-0.099}$ | 0.770 | 0.766 | 0.153 | 0.513 |
| F4 | $\frac{S}{y} = 0.05 * (\sigma)^{0.237}(Fr)^{0.231}(\frac{d50}{y})^{-0.162}(\frac{L}{y})^{0.553}$ | 0.629 | 0.966 | -0.086 | 0.599 |
| F5 | $\frac{S}{y} = 0.193 * (\sigma)^{0.222}(Fr)^{-0.012}(\frac{D}{y})^{0.191}(\frac{L}{y})^{0.202}$ | 0.658 | 1.018 | -0.054 | 0.629 |
| F6 | $\frac{S}{y} = 0.086 * (\sigma)^{0.081}(\frac{D}{y})^{0.193}(\frac{d50}{y})^{-0.110}(\frac{L}{y})^{0.352}$ | 0.722 | 0.849 | -0.018 | 0.570 |

Following Table 6, an inverse relationship between $\frac{d50}{y}$ and $\frac{S}{y}$ is found while the other variables including $\sigma, Fr, \frac{D}{y}, L/y$ have direct relationship with the target variable. Therefore, it is expected that pier foundation with finer sediments and larger width and length experience much weakening. Similarly, higher flow velocity and wider bed foundation materials (higher values of $\sigma$) associate with higher scour depth. Results of sensitivity analysis reveal that $D/y$ and $d50/y$ are amongst the most effective variables for pier scour in prototype scale. Performance of models *F5* and *F4* which exclude effect of these two variables demonstrate importance of them on the scour depth estimations. Moreover, it can be found that excluding Froude number degrade the model performance (*F6*) indicating effect of flow velocity on the scour process. On the other hand, excluding sediment gradation and pier length in the model development (*F2* and *F3*) does affect the model performance remarkably. Generally, the model (*F2*) including flow characteristics of depth and velocity (as Froude number), pier geometry (width and length), bed material specifications ($d50$) provides the most accurate estimations for scour depth. Estimations of the model (*F2*) have the highest correlation and the lowest errors compared to the other derived models. The width of uncertainty band for 95% confidence level for the derived equations were computed to provide more descriptions of reliability of estimations yielded by the equations. Figure 4 illustrates performance of the derived equations in terms of width of uncertainty band.

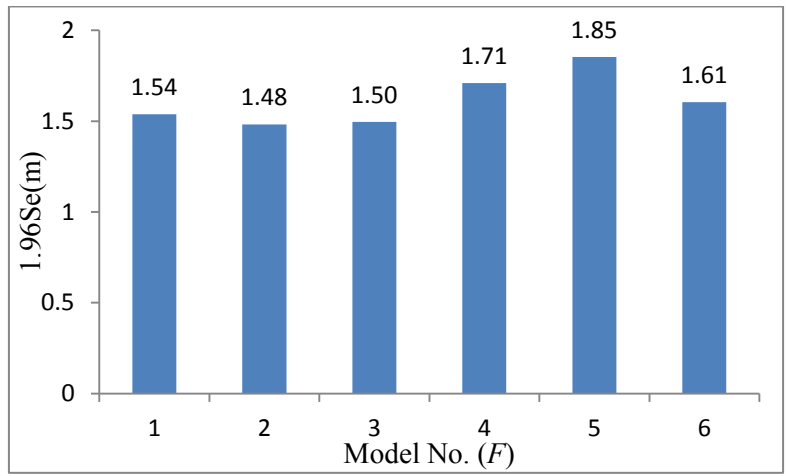

Figure 4. Width of uncertainty band for the derived equations (field measurements)

As observed from Figure 4, the first three models (*F1*, *F2*, and *F3*) have the narrowest uncertainty band compared with the others. Moreover, estimations of model *F2* expect to have the least uncertainty. Thus, this model is considered as the best model to estimate scour depth in field scale studies. However, efficiency of the model should be evaluated and compared with existing empirical equations to demonstrate its efficiency for real world applications. In this regard, performance of existing equations and also this model for testing dataset are presented in Table 7. Moreover, performance of the model *L6* proposed for laboratory dataset is presented in the table to assess its efficiency for scour depth prediction in prototype environment.

Table 7. Results of the best derived equations against existing equations for field measurements

| Model | $R^2$ | RMSE (m) | $\bar{e}$ (m) | MAE (m) | $1.96S_e$ (m) |
|---|---|---|---|---|---|
| Laursen and Toch (1956) | 0.540 | 3.898 | 2.802 | 2.802 | 5.310 |
| Shen et al. (1969) | 0.468 | 2.871 | 2.295 | 2.320 | 3.384 |
| Johnson (1992) | 0.336 | 1.372 | 0.284 | 0.881 | 2.631 |
| Richardson and | 0.624 | 2.226 | 1.839 | 1.893 | 2.459 |
| Davis (2001) | | | | | |
| HEC-18 (Mohamed et al., 2005) | 0.567 | 2.434 | 1.932 | 1.942 | 2.902 |
| Azamathulla et al. (2009) | 0.448 | 3.519 | 1.985 | 2.182 | 5.696 |
| Sharafi et al. (2016) | 0.710 | 0.869 | 0.123 | 0.607 | 1.687 |
| *L6* | 0.599 | 1.324 | 0.854 | 1.022 | 1.982 |
| *F2* | 0.774 | 0.753 | 0.030 | 0.520 | 1.474 |

Results of Table 7 imply that laboratory based equations are not efficient for scour depth estimation in prototype scale. High values of error and great amount of uncertainty are embedded with such equations when applied for filed applications. On the other hand, equations developed using field data provide relatively fair estimations of scour depth. Positive values of mean error shows that the models overestimate scour depth. Moreover, models such as *F2* and Sharafi et al. (2016) with high values of *MAE* and low values of mean error ($\bar{e}$), the errors are symmetrically distributed in which the models overestimations for some measurements are neutralized with underestimations for some other cases. Considering error indices, the model proposed in this study (*F2*) and the one developed by Sharafi et al. (2016) are among the most efficient models to predict scour depth in prototype applications. These two models have lower error values, narrower width of uncertainty and higher correlation with field measurements. However, model *F2* in which its coefficients were obtained using PSO optimization algorithm slightly outperforms the latter one in terms of *RMSE*, $R^2$, mean error and width of uncertainty band. Regardless these two equations, the model *L6* which developed using laboratory data outperforms the other equations in estimation of filed scale scour depth. This fact reflects capability of the proposed algorithm in obtaining nonlinear relationship of input and output and its superiority over previous linear and nonlinear regression models. Figure 5 illustrates scatter plot of estimated values and measured values of scour depth for testing dataset. It should be noticed that the testing dataset were randomly selected and excluded in PSO model developments.

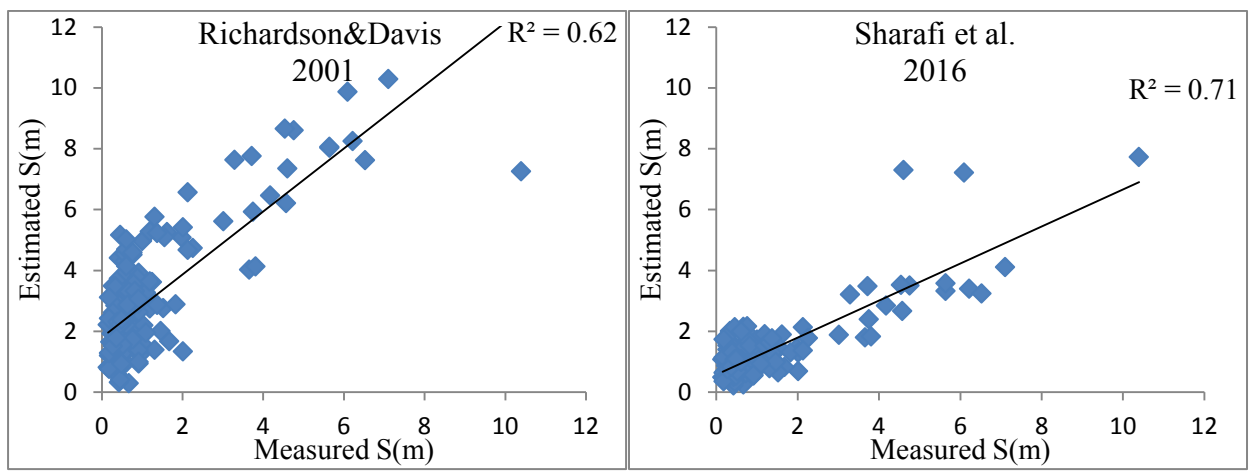

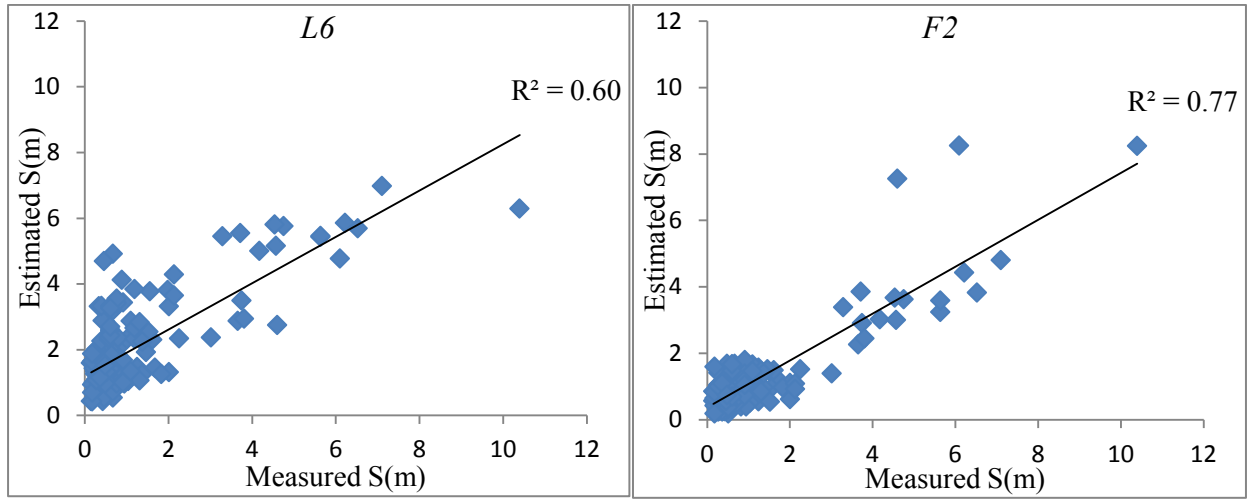

Figure 5. Scatter plot of estimated values versus field measurements of scour depth

Generally all the subplots indicate good agreement between estimated and measured values of scour depth. However, correlation between results of *F2* and observed values are stronger. The models provide sound predictions for low and medium ranges of scour depth (S<4m) while for extreme values (S>4m) all the models underestimate the target variable. This fact is in contrary to the positive values of mean error reported in Table 7 that reveals the models overestimate in most cases especially for low and medium ranges whereas for extreme values it is expected to have an underestimation.

Comparing coefficients of variables for laboratory and filed datasets, the constant coefficients represented by "$a$" in equations (7) and (8), it is change for a wider range when developed for laboratory dataset. In other words, it changes between [0-1] for field data while it has frequently larger values for laboratory dataset. Regarding exponential coefficient of $\sigma$, it has negative values for laboratory data indicating an inverse relationship between the variable with scour depth while for field data it takes positive value. This can illustrate difference between characteristics of laboratory and prototype models. Therefore, equations derived using laboratory data may not efficiently estimate scour depth for large scale applications. For the laboratory dataset, $\frac{V}{V_c}$, flow velocity and flow depth, pier width and sediment gradation and $d50$ were considered as effective parameters and for the field study the variable except $\frac{V}{V_c}$ that was replaced with $L/y$ were used. Therefore, to achieve a sound prediction of scour depth around bridge piers, different parameters describing flow characteristics, pier shape and geometry and also sediment and foundation material should be determined carefully. Pier length is an important parameter affecting scour phenomenon in real applications where it has been ignored in many laboratory studies. Therefore, its effect should be included in future studies to catch the physics of the phenomenon accordingly.

4. Conclusions

Reliable estimation of scour depth around bridge piers plays an important role in design of bridges and also for assessment of bridge safety. It is usually carried out using empirical equations based on a limited number of laboratory data. However, recently, some equations and also black box models were employed using field data. Empirical equations are cost effective due to requiring less computational efforts. Moreover, to a have single prediction for a bridge pier scours depth, a new model should be trained and constructed and datasets of previous measurements should be available for the model development. In case of using empirical equations, data are only needed for the case, but not for other previous measurements. Therefore, the empirical equations have advantages of easily implementations, less computational efforts, simplicity formulation and clearly relationship between input and target variables. However, efforts to improve efficiency and accuracy of empirical models are of great importance and scope of this study. In this regard, particle swarm optimization algorithm as a powerful tool to find optimized coefficients of empirical equations giving the least error of estimation was taken under consideration. Moreover, exploring and comparing efficiency of existing equations with those of proposed in this study was investigated. Finally, suitability of the equations developed using laboratory data for filed measurements was assessed.

This study provided empirical equations using extensive data of laboratory and field measurements. To do that, an optimization algorithm was employed to find optimum coefficients of the equations in which their general form was determined using available dimensional analysis. Performance of the derived equations for laboratory and field datasets was evaluated individually. Furthermore, sensitivity analysis was conducted to find the most effective parameters and the best combination of input variables for scour depth estimation. Main findings of the study can be summarized as the following points.

- The proposed equations for laboratory and filed scour depth investigations outperform existing equations considering error measures indices and uncertainty of estimations.
- In the laboratory based equations, pier length was ignored in the model development while the equation providing the most accurate predictions in prototype scale need the pier length as the model input in addition to other input variables.
- To achieve sound estimations of scour depth for large scale, equations only derived from laboratory datasets associate with high values of error and uncertainty.
- The optimization algorithm employed in this study is superior over previous regression based models. It has a great capability to catch nonlinear relationship between input variables and target variable.
- Considering results of sensitivity analysis, ration of pier width to flow depth and also sediment gradation were recognized as the most effective parameters for scour depth process in experimental scale while for large scale (field), ration of pier width to flow depth and the ratio of median particle size ($d50$) to flow depth were found to be the most effective variables.
- Positive values in exponential coefficient in the laboratory based data reveal direct relationships between Froude numbers, pier width and scour depth. On the other hand, inverse relationships for sediment gradation indicator, median particle size and scour depth were detected using the optimization algorithm.
- Similar to laboratory equation, exponential coefficients for Froude number, pier with, pier length had positive values. Also, negative value of the coefficient for median particle size describes an inverse relationship with scour depth. However, the coefficient for sediment gradation indicator for filed data showed a direct relationship that can reflect with increasing the indicator, scour depth is expected to increase. This is reflecting that in real filed applications, the sediment gradation has much variation and including much different materials associating with higher values of depth scour.

- The derived equations using the optimization algorithm provide reliable estimations for scour depth both in small and large scales. Moreover, the coefficients describing relationship between input and outputs variables are consistent with the physical concept of the phenomenon while in some previous equations the coefficient had variable signs incompatible with the physics of the process. This inconsistency may indicate limited number of the data was applied for the model development.

Finally, results of this study and the proposed equations derived using extensive datasets can be efficiently used to estimate scour depth around bridge piers. The equations provide more accurate estimations of the target value with lower amount of uncertainty. Using an optimization algorithm with great capability to recognize nonlinear relationship among variables, employing a large number of dataset, developing individual models for laboratory and field studies can be mentioned as main advantages of the study. The proposed method provides reliable and sound estimations while requires less computational efforts and much simplicity in formulation and application.

**References**

Alizadeh, M.J., Ahmadyar, D., Afghantoloee, A., 2017. Improvement on the existing equations for predicting longitudinal dispersion coefficient. Water resources management 31, 1777-1794.
Azamathulla, H.M., Ghani, A.A., 2010. ANFIS-based approach for predicting the scour depth at culvert outlets. Journal of pipeline systems engineering and practice 2, 35-40.
Azamathulla, H.M., Ghani, A.A., Zakaria, N.A., Guven, A., 2009. Genetic programming to predict bridge pier scour. Journal of Hydraulic Engineering 136, 165-169.
Bateni, S.M., Borghei, S., Jeng, D.-S., 2007. Neural network and neuro-fuzzy assessments for scour depth around bridge piers. Engineering Applications of Artificial Intelligence 20, 401-414.
Benedict, S.T., Caldwell, A.W., 2014. A pier-scour database: 2,427 field and laboratory measurements of pier scour. US Geological Survey Data Series 845.
Eberhart, R., Kennedy, J., 1995. A new optimizer using particle swarm theory, MHS'95. Proceedings of the Sixth International Symposium on Micro Machine and Human Science. Ieee, pp. 39-43.
Hancu, S., 1971. Sur le calcul des affouillements locaux dams la zone des piles des ponts, Proceedings of the 14th IAHR Congress, Paris, France, pp. 299-313.
Johnson, P., Clopper, P., Zevenbergen, L., Lagasse, P., 2015. Quantifying uncertainty and reliability in bridge scour estimations. Journal of Hydraulic Engineering 141, 04015013.
Johnson, P.A., 1992. Reliability-based pier scour engineering. Journal of Hydraulic engineering 118, 1344-1358.
Laursen, E.M., Toch, A., 1956. Scour around bridge piers and abutments. Iowa Highway Research Board Ames, IA.
Liao, K.-W., Lu, H.-J., Wang, C.-Y., 2015. A probabilistic evaluation of pier-scour potential in the Gaoping River Basin of Taiwan. Journal of Civil Engineering and Management 21, 637-653.
Melville, B., Sutherland, A., 1988. Design method for local scour at bridge piers. Journal of Hydraulic Engineering 114, 1210-1226.
Melville, B.W., Chiew, Y.-M., 1999. Time scale for local scour at bridge piers. Journal of Hydraulic Engineering 125, 59-65.
Mohamed, T.A., Noor, M., Ghazali, A.H., Huat, B.B., 2005. Validation of some bridge pier scour formulae using field and laboratory data. American Journal of Environmental Sciences 1, 119-125.

Mohamed, T.A., Pillai, S., Noor, M.J.M.M., Ghazali, A.H., Huat, B., Yusuf, B., 2006. Validation of some bridge pier scour formulae and models using field data. Journal of King Saud University-Engineering Sciences 19, 31-40.

Pal, M., Singh, N., Tiwari, N., 2012. M5 model tree for pier scour prediction using field dataset. KSCE Journal of Civil Engineering 16, 1079-1084.

Richardson, E., Davis, S., 2001. Evaluating Scour at Bridges: Hydraulic Engineering Circular No. 18. Rep. FHwA NHI, 01-001.

Sharafi, H., Ebtehaj, I., Bonakdari, H., Zaji, A.H., 2016. Design of a support vector machine with different kernel functions to predict scour depth around bridge piers. Natural Hazards 84, 2145-2162.

Shen, H.W., Schneider, V.R., Karaki, S., 1969. Local scour around bridge piers. Journal of the Hydraulics Division.

Tinoco, R., Goldstein, E., Coco, G., 2015. A data-driven approach to develop physically sound predictors: Application to depth-averaged velocities on flows through submerged arrays of rigid cylinders. Water Resources Research 51, 1247-1263.

Zounemat-Kermani, M., Beheshti, A.-A., Ataie-Ashtiani, B., Sabbagh-Yazdi, S.-R., 2009. Estimation of current-induced scour depth around pile groups using neural network and adaptive neuro-fuzzy inference system. Applied Soft Computing 9, 746-755.

Mosavi, A.*, P. Ozturk and K. W. Chau (2018): Flood prediction using machine learning models. Published in: Water 10(11).

Choubin, B.*, E. Moradi, M. Golshan, J. Adamowski, F. Sajedi-Hosseini and A. Mosavi (2019): An ensemble prediction of flood susceptibility using multivariate discriminant analysis, classification and regression trees, and support vector machines. Published in: Science of the Total Environment 651: 2087-2096.

Torabi, M., S. Hashemi, M. R. Saybani, S. Shamshirband* and A. Mosavi (2019): A Hybrid clustering and classification technique for forecasting short-term energy consumption. Published in: Environmental Progress and Sustainable Energy 38(1): 66-76.

Ardabili, S. F., B. Najafi, M. Alizamir, A. Mosavi, S. Shamshirband* and T. Rabczuk (2018): Using SVM-RSM and ELM-RSM approaches for optimizing the production process of methyl and ethyl esters. Published in: Energies 11(11).

Moeini, I., M. Ahmadpour, A. Mosavi, N. Alharbi and N. E. Gorji*(2018): Modeling the time-dependent characteristics of perovskite solar cells. Published in: Solar Energy 170: 969-973.

Najafi, B., S. F. Ardabili, A. Mosavi, S. Shamshirband* and T. Rabczuk (2018): An intelligent artificial neural network-response surface methodology method for accessing the optimum biodiesel and diesel fuel blending conditions in a diesel engine from the viewpoint of exergy and energy analysis. Published in: Energies 11(4).

Nosratabadi, S., A. Mosavi*, S. Shamshirband*, E. K. Zavadskas, A. Rakotonirainy, K.Wing-Chau (2019) Sustainable business models: a review. Published in: Sustainability 11(6), p.1663.

Mosavi, A., Salimi, M., Faizollahzadeh Ardabili, S., Rabczuk, T., Shamshirband, S. and Varkonyi-Koczy, A.R. (2019). State of the Art of Machine Learning Models in Energy Systems, a Systematic Review. Published in: Energies, 12(7), p.1301.

Qasem, S.N., Samadianfard, S., Nahand, H.S., Mosavi, A., Shamshirband, S*. and Chau, K.W. (2019). Estimating Daily Dew Point Temperature Using Machine Learning Algorithms. Published in: Water, 11(3), p.582.

Dineva, A., Mosavi, A., Ardabili, S.F., Vajda, I., Shamshirband, S.*, Rabczuk, T. and Chau, K.W. (2019). Review of soft computing models in design and control of rotating electrical machines. Published in: Energies, 12(6), p.1049.

Mosavi, A.*, (2019): Prediction of remaining service life of pavement using an optimized support vector machine. Published in: Engineering Applications of Computational Fluid Mechanics 13(1): 188-198.